%% file: main.tex
\documentclass[10pt,twocolumn,letterpaper]{article}

\usepackage[pagenumbers]{cvpr} 

\usepackage{amsfonts}
\usepackage{amsmath}
\usepackage{booktabs}
\usepackage{multirow}

\definecolor{cvprblue}{rgb}{0.21,0.49,0.74}
\usepackage[pagebackref,breaklinks,colorlinks,allcolors=cvprblue]{hyperref}

\def\paperID{} 
\def\confName{CVPR}
\def\confYear{2026}

\title{CLCR:Cross-Level Semantic Collaborative Representation for Multimodal Learning}

\author{Chunlei Meng\thanks{This study has been Accepted by CVPR 2026.}\thanks{The camera-ready version can be obtained from the official CVPR 2026 website.}\\
Fudan University\\
{\tt\small clmeng23@m.fudan.edu.cn}
\and
Guanhong Huang\\
Shantou University\\
\and
Rong Fu\\
University of Macau\\
\and
Runmin.JIAN\\
Guangzhou Huashang College\\
\and
Zhongxue Gan\\
Fudan University\\
\and
Chun Ouyang\thanks{Corresponding Author}\\
Fudan University\\
}

\begin{document}
\maketitle
\input{sec/0_abstract}

\input{sec/1_intro}

\input{sec/2_method}

\input{sec/3_experiment}

\input{sec/4_conclusion}

{
    \small
    \bibliographystyle{ieeenat_fullname.bst}
    \bibliography{main}
}

\input{sec/X_suppl}

\end{document}

%% file: sec/0_abstract.tex
\begin{abstract}
Multimodal learning aims to capture both shared and private information from multiple modalities. However, existing methods that project all modalities into a single latent space for fusion often overlook the asynchronous, multi-level semantic structure of multimodal data. This oversight induces semantic misalignment and error propagation, thereby degrading representation quality. To address this issue, we propose Cross-Level Co-Representation (CLCR), which explicitly organizes each modality's features into a three-level semantic hierarchy and specifies level-wise constraints for cross-modal interactions. First, a semantic hierarchy encoder aligns shallow, mid, and deep features across modalities, establishing a common basis for interaction. And then, at each level, an Intra-Level Co-Exchange Domain (IntraCED) factorizes features into shared and private subspaces and restricts cross-modal attention to the shared subspace via a learnable token budget. This design ensures that only shared semantics are exchanged and prevents leakage from private channels. To integrate information across levels, the Inter-Level Co-Aggregation Domain (InterCAD) synchronizes semantic scales using learned anchors, selectively fuses the shared representations, and gates private cues to form a compact task representation. We further introduce regularization terms to enforce separation of shared and private features and to minimize cross-level interference. Experiments on six benchmarks spanning emotion recognition, event localization, sentiment analysis, and action recognition show that CLCR achieves strong performance and generalizes well across tasks.
\end{abstract}

%% file: sec/1_intro.tex
\begin{figure}[htbp]
\centering
\includegraphics[width=1\linewidth]{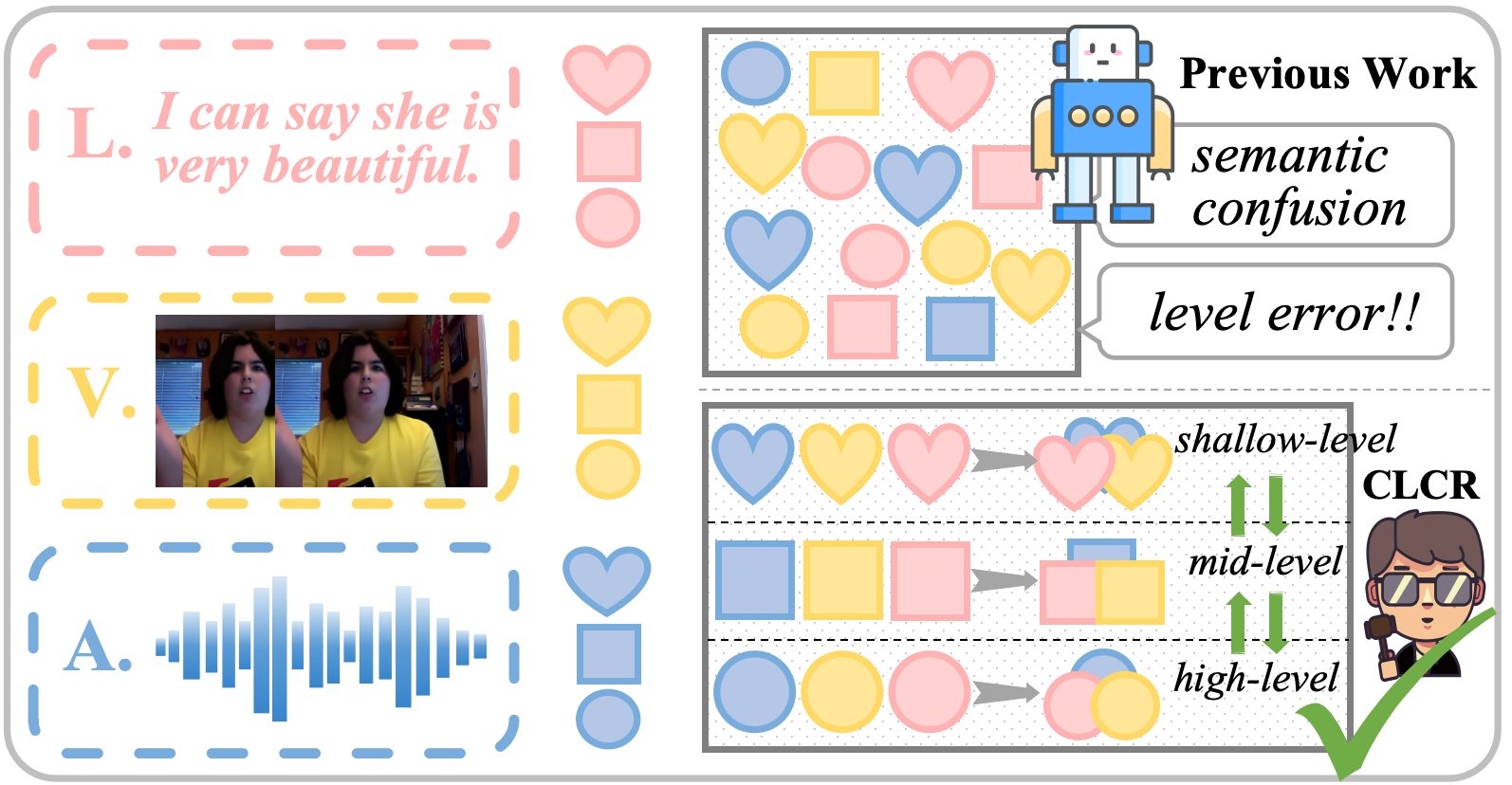}
\caption{Cross-level semantic asynchrony and CLCR. Top: mixing across levels causes semantic confusion and mismatch. bottom: CLCR structures each modality into three aligned levels and restricts exchange to the learned shared subspace at matched levels, enabling level semantic alignment and reliable aggregation.}
\label{fig:abs}
\end{figure}

\section{Introduction}
\label{sec:intro}

Multimodal learning (MML) aims to integrate information from multiple modalities (e.g., linguistic, visual, and acoustic) to obtain more comprehensive representations~\cite{ARL,Semi-IIN,cta-net,TSDA}. By combining complementary cues from different signal sources, multimodal systems can achieve more accurate and robust understanding than single-modal counterparts. MML has become central to applications such as human–computer interaction, sentiment analysis, and event understanding~\cite{misa,rts-vit,MENG-MGJR}. However, different modalities exhibit substantial discrepancies both in their semantic spaces and in the granularity of semantics across depth, which together pose significant challenges to current MML methods~\cite{FDMER,dmd,CF-ViT}.

To mitigate these challenges, existing research mainly follows two lines. The first line is based on feature disentanglement, which learns shared and modality-specific latent subspaces for multimodal representations~\cite{misa,FDMER,DLF}. MISA~\cite{misa} projects each modality into two subspaces (modality-invariant and modality-specific) and uses similarity losses and orthogonality constraints to enforce alignment in the common space and separation between common and private factors. FDMER~\cite{FDMER} employs encoders that extract common and private features separately and adopts an adversarial modality discriminator to further align the distributions of different modalities. DMD~\cite{dmd} performs graph-based cross-modal knowledge distillation in the decoupled subspaces to alleviate modality distribution mismatch, and DLF~\cite{DLF} designs geometric metrics to regularize the decoupling process. The second line dynamically calibrates modality contributions at both the sample and modality levels~\cite{MLA,rts-vit,d2r}. Typical mechanisms include mixture-of-experts weighting, cross-modal distillation, and imbalance-aware recalibration. For instance, EMOE~\cite{EMOE} adopts a modality-expert mixture, ARL~\cite{ARL} introduces a dual-path calibration strategy, DEVA~\cite{DEVA} converts audiovisual cues into text descriptions to leverage language models, and MLA~\cite{MLA} reformulates joint multimodal learning as an alternating single-modal training process. (More related Works, see Appendix.)

Although these methods alleviate modality heterogeneity and improve fusion, they generally assume that cross-modal interaction occurs at a single semantic level. In real-world data, however, evidence is organized hierarchically: shallow layers capture lexical or frame-level cues, middle layers encode phrases or prosodic structures, and deep layers reflect discourse intent or event context. As illustrated in Fig.~\ref{fig:abs}, when tokens from different levels are mixed without control, early fusion can induce semantic confusion and error propagation, cause private factors to leak into shared channels, and enforce excessive invariance that suppresses modality-specific cues needed for the task. From an information-theoretic perspective, let $Z$ denote the fused representation, $Y$ the task label, and $N$ nuisance factors. In an information bottleneck view, unstructured mixing of mismatched semantic levels tends to increase $I(Z;N)$ more than $I(Z;Y)$ under fixed capacity or invariance constraints, making it harder for downstream predictors to recover task-relevant information~\cite{tishby2000information,IB}. This cross-level semantic asynchrony, unlike generic modality heterogeneity, has been shown to degrade hierarchical multimodal modeling when left unmodeled~\cite{Han2021,Zhang2021}, and our analysis suggests that it is a key contributor to representation fragility. We explicitly target this factor with CLCR and empirically examine its impact through ablations in Section~\ref{sec:ablation}.

We address this gap with \textbf{Cross-Level Co-Representation (CLCR)}. CLCR first builds a three-level semantic hierarchy for each modality under a shared feature width, and confines cross-modal exchange at each level to an explicitly learned shared subspace under a token-level budget (IntraCED). This design limits mismatch propagation and prevents private-to-shared leakage at the source. It then synchronizes levels through anchor-based selection and performs cross-level aggregation without increasing representation dimensionality (InterCAD), while routing private summaries directly to the task heads to avoid cross-level mixing on the private path. Finally, intra- and inter-level regularizers preserve shared–private separation and penalize mixtures over incompatible level anchors, stabilizing level selection across modalities and depths. The architecture is trimodal by default yet applies to bimodal settings without structural changes. Across six benchmarks spanning emotion recognition, event localization, sentiment analysis, and action recognition, CLCR achieves high accuracy and exhibits robust performance while maintaining interpretable modality contributions. Overall, the main contributions of this work are as follows:
\begin{itemize}
  \item We propose \textbf{CLCR}, which organizes each modality into a three-level semantic hierarchy and explicitly specifies exchange and alignment rules to handle cross-level semantic heterogeneity.
  \item We introduce \textbf{IntraCED} and \textbf{InterCAD}: IntraCED performs budgeted shared-only token exchange at each level, and InterCAD provides anchor-guided cross-level aggregation with private routing, thereby reducing mismatch propagation and preserving modality-specific cues.
  \item We design intra- and inter-level regularization losses that stabilize shared–private separation and level selection.
\end{itemize}

%% file: sec/2_method.tex
\begin{figure*}[ht]
\centering
\includegraphics[width=1\linewidth]{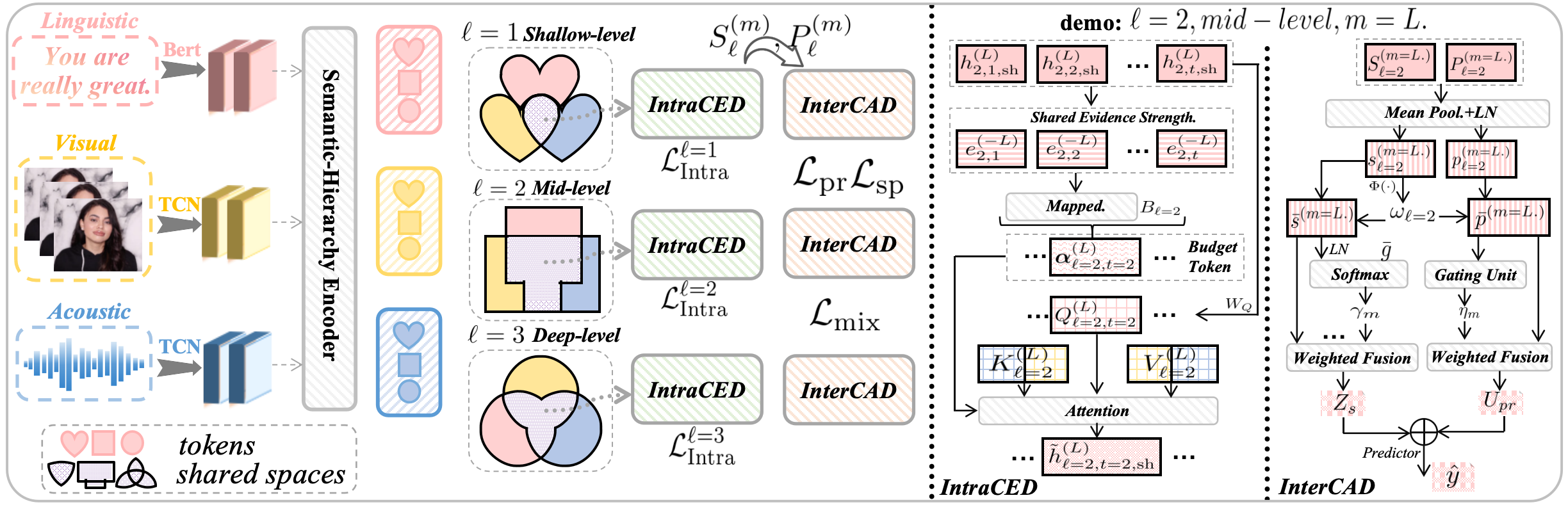}
\caption{Overview of the proposed CLCR framework. Each modality is organized into a three-level semantic hierarchy. IntraCED performs budgeted cross-modal exchange in an explicitly disentangled level-shared subspace, while InterCAD synchronizes and aggregates shared and private streams across levels into the final task representation.}
\label{fig:CLCR}
\end{figure*}

\section{Method}

\subsection{Model Overview}
The CLCR framework addresses cross-level semantic asynchrony in multimodal learning by organizing each modality into a three-level hierarchy and restricting cross-modal exchange to an explicitly disentangled level-shared subspace. As shown in Fig.~\ref{fig:CLCR}, a semantic-hierarchy encoder produces shallow, mid, and deep-level token sequences with a common feature width; IntraCED factorizes each level into modality-shared and modality-private subspaces and allows budgeted cross-modal exchange only in the shared subspace; and InterCAD summarizes shared and private streams across levels, synchronizes semantic scales, performs modality selection on the shared path, and aggregates private information through confidence gating. Two regularizers maintain shared–private separation across levels and modalities and, together with the task loss, define the overall training objective.

\subsection{Semantic-Hierarchy Encoder}
We consider linguistic, visual, and acoustic modalities. For each modality \(m \in \{\text{L},\text{V},\text{A}\}\), the input is a sequence $X^{(m)} = \{x^{(m)}_t\}_{t=1}^{T^{(m)}}, \quad x^{(m)}_t \in \mathbb{R}^{d_0}.$ To mitigate cross-level semantic asynchrony, we build a three-level hierarchy \(\ell \in \{1,2,3\}\) that captures shallow, mid, and deep semantics. This design is consistent with multi-scale representation studies showing that combining early, intermediate, and late features captures most of the useful variation in deep encoders~\cite{2025-low}. For each modality and level, the backbone output is mapped to a unified feature space of width d and augmented with positional encodings and layer normalization:  $H^{(m)}_{\ell}=\mathrm{LN}\big(Z^{(m)}_{\ell} W^{(m)}_{\ell} + P^{(m)}_{\ell}\big)$, where \(Z^{(m)}_{\ell}\) is the backbone output at level \(\ell\), \(W^{(m)}_{\ell} \in \mathbb{R}^{d_m \times d}\) is a linear projection with \(d_m \in \{d_{\text{bert}}, d_{\text{tcn}}\}\), and \(P^{(m)}_{\ell}\) encodes order within each level. This yields level-wise features \(H^{(m)}_{\ell} \in \mathbb{R}^{T^{(m)}_{\ell} \times d}\). Positional encodings are defined independently within each modality and level, so that local temporal order is preserved without assuming framewise alignment across modalities.

The linguistic modality is encoded by a pretrained BERT~\cite{MAG-BERT}, whose early, middle, and late layers provide shallow, mid, and deep-level features. These three levels respectively focus on lexical and syntactic patterns (subword structure, part-of-speech cues), phrasal and clause-level compositional and sentiment cues, and discourse-level intent with long-range dependencies. The visual and acoustic streams are encoded by three-stage Temporal Convolutional Networks (TCN)~\cite{TCN} with increasing receptive fields. In the visual branch, the three levels capture local appearance and motion primitives, part-level structure and short actions, and long-range scene and event context. In the acoustic branch, they capture frame-level spectra and micro-prosody, phoneme and syllable patterns with timbre dynamics, utterance-level prosody, and emotional contours. The encoder outputs the collection \(\{H^{(m)}_{\ell}\}_{m,\ell}\), which is level aligned in semantics and width aligned in channels. These features serve as inputs to IntraCED and InterCAD.

\subsection{Intra-Level Co-Exchange Domain (IntraCED)}
Existing methods often mix tokens from different semantic levels, which produces mismatched interactions and allows modality-specific factors to leak into the shared interaction channels. To address this, IntraCED is applied independently at each semantic level \(\ell\) and follows three principles: separate modality-invariant content into a shared subspace, restrict cross-modal attention to that subspace, and limit the exchanged information with a token-level budget. Each level becomes a controlled exchange domain that produces level-synchronous shared signals for InterCAD while preserving modality-specific private streams.

Formally, given level-wise representations $H^{(m)}_{\ell}=\{h^{(m)}_{\ell,t}\}_{t=1}^{T^{(m)}_{\ell}}\in\mathbb{R}^{T^{(m)}_{\ell}\times d}$ from modality $m$ at level $\ell$, IntraCED constructs an orthogonal shared–private decomposition. For each level $\ell$, we learn two projectors:
\begin{equation}
P^{\mathrm{sh}}_{\ell}=U^{\mathrm{sh}}_{\ell}(U^{\mathrm{sh}}_{\ell})^{\top}, \qquad
P^{\mathrm{pr}}_{\ell,m}=U^{\mathrm{pr}}_{\ell,m}(U^{\mathrm{pr}}_{\ell,m})^{\top}
\label{eq:intra-shared-proj}
\end{equation}
where $U^{\mathrm{sh}}_{\ell}\in\mathbb{R}^{d\times k}$ and $U^{\mathrm{pr}}_{\ell,m}\in\mathbb{R}^{d\times(d-k)}$ are learned orthonormal bases with $(U^{\mathrm{sh}}_{\ell})^{\top}U^{\mathrm{pr}}_{\ell,m}=0$. We enforce orthonormality during training via a Stiefel parameterization. Each token \(h^{(m)}_{\ell,t} \in \mathbb{R}^{1 \times d}\) is decomposed as:
\begin{equation}
h^{(m)}_{\ell,t,\mathrm{sh}} = h^{(m)}_{\ell,t} P^{\mathrm{sh}}_{\ell},
\qquad
h^{(m)}_{\ell,t,\mathrm{pr}} = h^{(m)}_{\ell,t} P^{\mathrm{pr}}_{\ell,m},
\label{eq:intra-split}
\end{equation}
so that \(h^{(m)}_{\ell,t}=h^{(m)}_{\ell,t,\mathrm{sh}}+h^{(m)}_{\ell,t,\mathrm{pr}}\). Only the shared component is exposed to cross-modal exchange, while the private remains isolated and is routed along the private path.

\textbf{Budgeted token selection.} Not all shared tokens are equally reliable carriers of cross-modal evidence. To avoid dense and noisy fusion, IntraCED assigns a level-wise budget that limits how many tokens can participate in exchange. For each token, we measure its shared evidence strength $e^{(m)}_{\ell,t} = \|h^{(m)}_{\ell,t,\mathrm{sh}}\|_{2}.$ This strength is mapped to a preliminary activation weight $\tilde\alpha^{(m)}_{\ell,t} = \sigma\big(\rho\, e^{(m)}_{\ell,t} - \theta_{\ell}\big),$ where \(\rho > 0\) is a learnable scale and \(\theta_{\ell}\) is a level-specific threshold. Collecting all token weights yields \(\tilde{\boldsymbol{\alpha}}^{(m)}_{\ell} = [\tilde\alpha^{(m)}_{\ell,1}, \dots, \tilde\alpha^{(m)}_{\ell,T^{(m)}_{\ell}}]^{\top}\). We then project this vector onto a truncated simplex:
\begin{equation}
\begin{split}
\boldsymbol{\alpha}^{(m)}_{\ell}
&= \operatorname{Proj}_{\Delta(B_{\ell})}\big(\tilde{\boldsymbol{\alpha}}^{(m)}_{\ell}\big), \\
\Delta(B_{\ell})
&= \Big\{\mathbf{a} \in [0,1]^{T^{(m)}_{\ell}} : \mathbf{1}^{\top} \mathbf{a} \le B_{\ell} \Big\},
\end{split}
\label{eq:intra-budget}
\end{equation}
where \(B_{\ell}\) is a learnable budget at level \(\ell\). This projection produces a sparse, capacity-limited set of tokens that are allowed to receive cross-modal information.

\textbf{Tri-modal level-shared space exchange.} At each level, a target modality aggregates information from the shared components of the remaining modalities. Let \(\mathcal{M}\) be the modality set and define \(\mathcal{S}(m) = \mathcal{M} \setminus \{m\}\). We stack shared tokens from sources other than modality at \(\ell\) layer, as $H^{(-m)}_{\ell,\mathrm{sh}}
=
\operatorname{Concat}\big(\{H^{(n)}_{\ell,\mathrm{sh}}\}_{n \in \mathcal{S}(m)}\big)
\in \mathbb{R}^{T^{(-m)}_{\ell} \times d},$ where \(T^{(-m)}_{\ell} = \sum_{n \in \mathcal{S}(m)} T^{(n)}_{\ell}\) and \(H^{(n)}_{\ell,\mathrm{sh}}\) stacks \(\{h^{(n)}_{\ell,t,\mathrm{sh}}\}_t\). Query, key, and value projections are
$Q^{(m)}_{\ell,t} = h^{(m)}_{\ell,t,\mathrm{sh}} W_{Q},$ $K^{(-m)}_{\ell} = H^{(-m)}_{\ell,\mathrm{sh}} W_{K},$ $V^{(-m)}_{\ell} = H^{(-m)}_{\ell,\mathrm{sh}} W_{V},$ with \(W_{Q}, W_{K}, W_{V} \in \mathbb{R}^{d \times d_h}\). Cross-modal attention in the shared subspace is $\tilde h^{(m)}_{\ell,t,\mathrm{sh}}$:
\begin{equation}
\tilde h^{(m)}_{\ell,t,\mathrm{sh}}=\alpha^{(m)}_{\ell,t}\,
\mathrm{Attn}\big(Q^{(m)}_{\ell,t}, K^{(-m)}_{\ell}, V^{(-m)}_{\ell}\big)
\end{equation}
Thus each modality queries a joint shared pool formed by the others, while the budget $\alpha^{(m)}_{\ell,t}$ gates how much external evidence each token absorbs. In bimodal settings, \(\mathcal{S}(m)\) has a single source modality and the formulation reduces to standard cross-attention.

\textbf{Intra-Level Regularization $\mathcal{L}_{\mathrm{Intra}}$.} To keep private and shared streams statistically distinct, IntraCED uses an identifiability regularizer based on whitened cross-correlation. Let $S^{(m)}{\ell}$ and $P^{(m)}{\ell}$ denote the updated shared $S^{(m)}_{\ell}$ and private $P^{(m)}_{\ell}$ streams, each as a matrix in \(\mathbb{R}^{T^{(m)}_{\ell} \times d}\).
\begin{equation}
S^{(m)}_{\ell} = \{\tilde h^{(m)}_{\ell,t,\mathrm{sh}}\}_t,
\qquad
P^{(m)}_{\ell} = \{h^{(m)}_{\ell,t,\mathrm{pr}}\}_t
\end{equation}

Given matrices \(A \in \mathbb{R}^{T_A \times d}\) and \(B \in \mathbb{R}^{T_B \times d}\), we center rows and estimate sample covariances \(\hat\Sigma_A\), \(\hat\Sigma_B\) and cross-covariance \(\hat\Sigma_{AB}\). The whitened correlation operator is
\begin{equation}
\mathrm{Corr}(A,B)
=
\hat\Sigma_A^{-\frac{1}{2}} \hat\Sigma_{AB} \hat\Sigma_B^{-\frac{1}{2}},
\end{equation}
with \(\hat\Sigma_A^{-\frac{1}{2}}\) and \(\hat\Sigma_B^{-\frac{1}{2}}\) obtained by eigendecomposition with diagonal regularization. 
From a computational perspective, both the Stiefel parameterization of $U^{\mathrm{sh}}_{\ell}$ and $U^{\mathrm{pr}}_{\ell,m}$ and the whitening-based $\mathrm{Corr}(\cdot,\cdot)$ operator act on $d \times d$ statistics at each level, so the overhead is negligible compared to the backbone encoders and attention layers. The intra-level identifiability loss:
\begin{equation}
\begin{split}
\mathcal{L}_{\mathrm{Intra}}
\sum_{\ell}
\Big(
\sum_{m\neq n}\|\mathrm{Corr}(P^{(m)}_{\ell},P^{(n)}_{\ell})\|_{F}^{2}
+ \\
\lambda_{\mathrm{sp}}\sum_{m}\|\mathrm{Corr}(P^{(m)}_{\ell},S^{(m)}_{\ell})\|_{F}^{2}
\Big)
\label{eq:intra-id}
\end{split}
\end{equation}

where \(\lambda_{\mathrm{sp}} \in (0,1]\) controls the separation strength. The first term discourages different modalities from encoding similar content in their private subspaces at the same level. The second term discourages leakage between private and shared components within each modality. Combined with the shared–private projections and budgeted token gating, this loss keeps the shared space focused on cross-modal evidence and preserves modality-specific information in the private streams.

\subsection{Inter-Level Co-Aggregation Domain}

After IntraCED, each modality \(m\) and level \(\ell\) has an updated shared stream \(S^{(m)}_{\ell}\) and a private stream \(P^{(m)}_{\ell}\). 


\textbf{Cross-level semantic synchronization.} For each modality and level, InterCAD first compresses shared and private streams into fixed-size summaries \(s^{(m)}_{\ell}\) and \(p^{(m)}_{\ell}\) using mean pooling followed by layer normalization. We obtain hierarchical anchors \(g_{\ell}\) by averaging and normalizing the shared summaries across modalities at level \(\ell\). The three anchors are concatenated and passed through a two-layer perceptron \(\Phi(\cdot)\), followed by a softmax to obtain the level weights:
\begin{equation}
\omega = \mathrm{softmax}\big(\Phi([g_1; g_2; g_3])\big) \in \Delta^{3},
\label{eq:InterCAD-omega}
\end{equation}
so that \(\omega = [\omega_1,\omega_2,\omega_3]^{\top}\) lies on the probability simplex and assigns an importance score to each level. Shared and private summaries are synchronized across levels as
\begin{equation}
\bar{s}^{(m)} = \sum_{\ell=1}^{3} \omega_{\ell} s^{(m)}_{\ell},
\qquad
\bar{p}^{(m)} = \sum_{\ell=1}^{3} \omega_{\ell} p^{(m)}_{\ell}.
\end{equation}

\textbf{Modality Selection and Private Information Aggregation.} After performing cross-level semantic synchronization, InterCAD focuses on selecting the most informative modality for each task. We compute a global shared context $\bar{g}$ by averaging the synchronized shared summaries ${\bar{s}^{(m)}}$ across modalities. This context vector serves as the query for modality selection. We compute modality-specific keys \(k_m\) for each modality \(m\), and use these to calculate the attention weights \(\gamma\) through scaled dot-product attention. Where \(q = \mathrm{LN}(W_q \bar{g})\) is the query derived from the global shared context, and \(k_m = \mathrm{LN}(W_k \bar{s}^{(m)})\) represents the key for modality \(m\). These attention weights \(\gamma\) are then used to obtain the fused shared descriptor \(\bar{z}_{\mathrm{sh}}\), which is a weighted sum of the shared representations from each modality.
\begin{equation}
\gamma = \mathrm{softmax}\left( \frac{q [k_1, \dots, k_M]^{\top}}{\sqrt{d}} \right),z_{\mathrm{sh}} = \sum_{m=1}^{M} \gamma_m \bar{s}^{(m)}.
\end{equation}
For the aggregation of private information, we use a confidence gate for each modality. The confidence score \(\eta_m\) for modality \(m\) is computed based on the private stream \(\bar{p}^{(m)}\), and it reflects the reliability of the private representation. The final aggregated private descriptor \(u_{\mathrm{pr}}\) is then obtained by summing the weighted private representations:
\begin{equation}
\eta_m = \sigma\left( w_p^{\top} \mathrm{LN}(W_p \bar{p}^{(m)}) \right),
u_{\mathrm{pr}} = \sum_{m=1}^{M} \eta_m \bar{p}^{(m)}.
\end{equation}

\textbf{Task Representation.} We obtain the final task representation by concatenating the fused shared descriptor and the aggregated private descriptor and feeding the result into the task head $\hat{y} = f_{\theta}(z_{\mathrm{sh}} \oplus u_{\mathrm{pr}})$, where $f_{\theta}$ denotes a classifier, or regressor depending on the task.

\textbf{Inter-level Regularization $\mathcal{L}_{\mathrm{Inter}}$.} To enforce cross-level consistency and avoid asynchronous mixing, we introduce an inter-level regularizer with three correlation-based terms on private streams, shared–private interactions, and level selection.
Semantic incompatibility between two level anchors $g_{\ell}$ and $g_{k}$ is $\delta_{\ell k}
=
1 - \frac{1}{d} \|\mathrm{Corr}(g_{\ell}, g_{k})\|_{F}^{2},$ so similar anchors yield small penalties and dissimilar anchors large ones.
We define the three inter-level terms as
$\mathcal{L}_{\mathrm{pr}}
=
\sum_{m=1}^{M} \sum_{\ell \ne k}
\|\mathrm{Corr}(P^{(m)}_{\ell}, P^{(m)}_{k})\|_{F}^{2}$,
$\mathcal{L}_{\mathrm{sp}}
=
\sum_{m=1}^{M} \sum_{\ell,k}
\|\mathrm{Corr}(S^{(m)}_{\ell}, P^{(m)}_{k})\|_{F}^{2}$,
and
$\mathcal{L}_{\mathrm{mix}}
=
\sum_{\ell < k} \omega_{\ell} \omega_{k}\, \delta_{\ell k}$, where
$\mathcal{L}_{\mathrm{pr}}$ reduces private redundancy across depths,
$\mathcal{L}_{\mathrm{sp}}$ suppresses shared–private leakage, and
$\mathcal{L}_{\mathrm{mix}}$ discourages assigning mass to incompatible level pairs. The full loss is
\begin{equation}
\mathcal{L}_{\mathrm{Inter}}
=
\alpha_{\mathrm{pr}} \mathcal{L}_{\mathrm{pr}}
+
\alpha_{\mathrm{sp}} \mathcal{L}_{\mathrm{sp}}
+
\alpha_{\mathrm{mix}} \mathcal{L}_{\mathrm{mix}},
\label{eq:intercad-reg}
\end{equation}
with $\alpha_{\mathrm{pr}}, \alpha_{\mathrm{sp}}, \alpha_{\mathrm{mix}} > 0$.
This regularizer stabilizes level-synchronized summaries by removing redundant private information and penalizing incompatible level combinations.

\subsection{Objective Optimization}
CLCR is trained end-to-end with a standard task loss and two complementary regularizers. For classification tasks we use cross-entropy, and for regression tasks we use mean-squared error. In addition to the task loss \(\mathcal{L}_{\mathrm{task}}\), we apply the intra-level identifiability loss \(\mathcal{L}_{\mathrm{Intra}}\) in \eqref{eq:intra-id} and the inter-level regularization loss \(\mathcal{L}_{\mathrm{Inter}}\) in \eqref{eq:intercad-reg}. The overall objective is
\begin{equation}
\mathcal{L}_{\mathrm{all}}
=
\mathcal{L}_{\mathrm{task}}
+
\lambda_{\mathrm{inter}} \mathcal{L}_{\mathrm{Inter}}
+
\lambda_{\mathrm{intra}} \mathcal{L}_{\mathrm{Intra}},
\label{eq:overall-objective}
\end{equation}
where \(\lambda_{\mathrm{inter}}, \lambda_{\mathrm{intra}} > 0\) are trade-off weights tuned on validation data. This objective jointly encourages accurate task predictions, cross-level disentanglement, and preservation of modality-specific identity.

%% file: sec/3_experiment.tex
\begin{table*}[t]
\small
\caption{Comparison with state-of-the-art methods on the benchmarks (\%). Mean and standard deviation are reported over 5 folds.}
\centering 
\small
\setlength{\tabcolsep}{3pt} 
\renewcommand{\arraystretch}{0.9} 
\begin{tabular}{c||cc||cc||cc||cc} 

\toprule 
\multirow{2}{*}{\textbf{Methods}} & \multicolumn{2}{c||}{\textbf{CREMA-D~\cite{crema-D}}} & \multicolumn{2}{c||}{\textbf{KS~\cite{KS}}} & \multicolumn{2}{c||}{\textbf{AVE~\cite{ARL}}} & \multicolumn{2}{c}{\textbf{UCF101~\cite{ucf101}}} \\ & Acc(\%) & F1(\%) & Acc(\%) & F1(\%)  & Acc(\%) & F1(\%)  & Acc(\%) & F1(\%)  \\ 
\hline \hline 
only-Acoustic & 57.27 & 57.89 & 48.67 & 48.89 & 62.16 & 58.54 & - & - \\ 
only-Visual & 62.17 & 62.78 & 52.36 & 52.67 & 31.40 & 29.87 & - & - \\ 

Concat & 58.83 & 59.43 & 64.97 & 65.21 & 66.15 & 62.46 & 80.41 & 79.40 \\ 
Grad-Blending & 68.81 & 69.34 & 67.31 & 67.68 & 67.40 & 63.87 & 81.73 & 80.84 \\ 
OGM-GE~\cite{OGM} & 64.34 & 64.93 & 66.35 & 66.76 & 65.62 & 62.97 & 81.15 & 80.36 \\ 
AGM~\cite{AGM} & 67.21 & 68.04 & 65.61 & 65.99 & 64.50 & 61.49 & 81.55 & 80.36 \\ 
PMR~\cite{PMR} & 65.12 & 65.91 & 65.01 & 65.13 & 63.62 & 60.36 & 81.36 & 80.37 \\ 
MMPareto~\cite{mmpareto} & 70.19 & 70.82 & 69.13 & 69.05 & 68.22 & 64.54 & 81.98 & 80.64 \\ 
MLA~\cite{MLA} & 73.21 & 73.77 & 69.62 & 69.98 & 70.92 & 67.23 & 82.01 & 81.22 \\ 
D\&R~\cite{d2r} & 73.52 & 73.96 & 69.10 & 69.36 & 69.62 & 64.93 & 82.11 & 80.87 \\ 
ARL~\cite{ARL} & 76.46 & 76.85 & 74.09 & 74.01 & 72.61 & 67.98 & 83.06 & 81.78 \\ 
\hline 
\textbf{CLCR (Ours)} & \textbf{77.92{\scriptsize$\pm$0.08}} & \textbf{78.33{\scriptsize$\pm$0.07}} & \textbf{75.41{\scriptsize$\pm$0.10}} & \textbf{75.32{\scriptsize$\pm$0.09}} & \textbf{73.82{\scriptsize$\pm$0.12}} & \textbf{69.11{\scriptsize$\pm$0.11}} & \textbf{83.64{\scriptsize$\pm$0.05}} & \textbf{82.27{\scriptsize$\pm$0.06}} 
\\

\toprule 
\end{tabular} 
\label{tab:main-1}
\end{table*}

\section{Experiments}
\label{experiment}
\subsection{Experimental Settings}

\textbf{Benchmarks.} We evaluate CLCR on six multimodal benchmarks spanning emotion recognition, event localization, sentiment analysis, and human action recognition. CREMA-D~\cite{crema-D} is an audio–visual emotion dataset with 7,442 speech–face clips in six classes (6,698/744 train/test). AVE~\cite{ARL} contains 4,143 ten-second audio–visual clips in 28 event categories, following~\cite{ARL}, we use synchronized localized segments for multimodal classification. Kinetics-Sounds (KS)~\cite{KS} comprises about 19K ten-second clips of 34 actions with both visual and auditory cues (15K/1.9K/1.9K train/val/test). UCF101~\cite{ucf101} provides RGB and optical flow modalities for 101 actions, and we use the official split (9,537/3,783 train/test). CMU-MOSI~\cite{Cmu-mosi} contains 2{,}199 monologue video segments with acoustic and visual features (1,284/229/686 train/val/test) annotated on a $[-3,3]$ sentiment scale, and we adopt the standard binary positive/negative setting. CMU-MOSEI~\cite{Cmu-mosei} includes 22,856 segments (16,326/1,871/4,659 train/val/test) with sentiment labels in the same range. 

\textbf{Implementation Details.} We train CLCR using SGD with momentum 0.9, a learning rate of $1\times10^{-3}$, weight decay of $1\times10^{-4}$, a batch size of 64, and 100 epochs on A100 GPUs. For CREMA-D, AVE, and KS, we follow the preprocessing and training settings of~\cite{ARL}.On MOSI and MOSEI, we report MAE, ACC$_2$, ACC$_7$, and F1, on the other four benchmarks, we report accuracy and F1.

\begin{table*}[t]
\small
\caption{Performance Comparison. $\uparrow$ and $\downarrow$ indicate that higher and lower value is better. Mean and StDev. are reported over 5 folds.}
\centering
\setlength{\tabcolsep}{2pt} 
\renewcommand{\arraystretch}{0.9} 
\begin{tabular}{l||ccccc||ccccc}
\toprule
\multirow{2}{*}{\textbf{Methods}} 
& \multicolumn{5}{c||}{\textbf{CMU-MOSI}} 
& \multicolumn{5}{c}{\textbf{CMU-MOSEI}} \\
 & MAE ($\downarrow$) & Corr ($\uparrow$) & Acc-2(\%) & Acc-7(\%)& F1(\%)
 & MAE ($\downarrow$) & Corr ($\uparrow$) & Acc-2(\%) & Acc-7(\%) & F1(\%) \\
\midrule
\midrule
MulT~\cite{MuLT}& 0.846 & 0.725 & 81.70 & 40.05 & 81.66 & 0.673 & 0.677 & 80.85 & 48.37 & 80.86 \\
MISA~\cite{misa}     & 0.788 & 0.744 & 82.07 & 41.27 & 82.43 & 0.594 & 0.724 & 82.03 & 51.43 & 82.13 \\
Self-MM~\cite{self-mm}& 0.765 & 0.764 & 82.88 & 42.03 & 83.04 & 0.576 & 0.732 & 82.43 & 52.68 & 82.47 \\
FDMER~\cite{FDMER}& 0.760 & 0.777 & 83.01 & 42.88 & 83.22 & 0.571 & 0.743 & 83.88 & 53.21 & 83.35 \\
PMR~\cite{PMR}& 0.895 & 0.689 & 79.88 & 40.60 & 79.83 & 0.645 & 0.689 & 81.57 & 48.88 & 81.56 \\
DMD~\cite{dmd}& 0.744 & 0.788 & 83.24 & 43.88 & 83.55 & 0.561 & 0.758 & 84.17 & 54.18 & 83.88 \\
CGGM~\cite{CGGM}& 0.747 & 0.798 & 84.43 & 43.21 & 84.13 & 0.551 & 0.761 & 83.90 & 53.47 & 84.14 \\
DEVA~\cite{DEVA}& 0.730 & 0.787 & 84.40 & 46.33 & 84.45 & 0.541 & 0.769 & 83.17 & 52.28 & 83.75 \\
ARL~\cite{ARL}& - & - & 84.16 & - & 84.09 & - & - & 83.65 & - & 83.79 \\
DLF~\cite{DLF}& 0.731 & 0.781 & 85.06 & 47.08 &85.04& 0.536 & 0.764 & 85.42 & 53.9 & 85.27 \\
EMOE~\cite{EMOE}& 0.710 & - & 85.4 & 47.7 & 85.4 & 0.536 & - & 85.3 & 54.1 & 85.3 \\
\hline
\textbf{CLCR (Ours)}&
\textbf{0.678}{\textbf{\scriptsize$\pm$0.007}} & \textbf{0.819}{\textbf{\scriptsize$\pm$0.005}} & \textbf{88.05}{\textbf{\scriptsize$\pm$0.20}} & \textbf{49.56}{\textbf{\scriptsize$\pm$0.25}} & \textbf{87.99}{\textbf{\scriptsize$\pm$0.18}} &
\textbf{0.511}{\textbf{\scriptsize$\pm$0.006}} & \textbf{0.790}{\textbf{\scriptsize$\pm$0.004}} & \textbf{87.96}{\textbf{\scriptsize$\pm$0.21}} & \textbf{56.32}{\textbf{\scriptsize$\pm$0.27}} & \textbf{88.02}{\textbf{\scriptsize$\pm$0.19}} \\

\bottomrule
\end{tabular}
\label{tab:main-2}
\end{table*}

\subsection{Comparison with the State-of-the-Art}

We compare CLCR with recent baselines on multiple audio-visual and Multimodal Sentiment Analysis (MSA) benchmarks, using 5-fold cross-validation in all experiments.

\textbf{Results on Acoustic–Visual Task.} As shown in Table~\ref{tab:main-1}, on CREMA-D, KS, and AVE, CLCR achieves the best accuracy and F1 under the same five-fold cross-validation protocol. Compared to the strongest baseline, CLCR improves accuracy by 1.46\%, 1.32\%, and 1.21\% on CREMA-D, KS, and AVE, respectively, and improves F1 by 1.48\%, 1.31\%, and 1.13\%. On UCF101, CLCR further improves accuracy and F1 by 0.58\% and 0.49\% over the best. These gains are consistent across datasets with different temporal dynamics, indicating that CLCR generalizes beyond a single corpus.

\textbf{Results on MSA.} As summarized in Table~\ref{tab:main-2}, CLCR consistently reduces MAE and improves classification metrics on CMU-MOSI and CMU-MOSEI. Relative to the strongest baseline, CLCR reduces MAE by 0.032 on MOSI and by 0.025 on MOSEI, while improving Acc$_2$ by 2.65\% and 2.54\% and F1 by 2.59\% and 2.72\%, respectively. These results show that the cross-level design of CLCR also transfers well to regression-style sentiment prediction.

\textbf{CLCR advantages analysis.}
Compared with baseline that lack explicit cross-level structuring and with disentanglement designs that only decouple modalities at a single level, CLCR jointly addresses within-level and cross-level mismatch. It organizes each modality into aligned shallow, mid, and deep layers, factorizes each layer into shared and private subspaces, restricts cross-modal exchange to a budgeted shared subspace, and applies cross-level regularization to suppress private redundancy and shared–private leakage between incompatible levels. This combination yields more stable improvements over diverse baselines across datasets and metrics.

\begin{table}[ht]
\small
\caption{Ablation studies of CLCR on three benchmarks (\%). $\ddagger$ denotes removing both $\mathcal{L}_{\mathrm{Inter}}$ and $\mathcal{L}_{\mathrm{Intra}}$.}
\centering

\setlength{\tabcolsep}{2.6pt}
\renewcommand{\arraystretch}{0.90}
\begin{tabular}{l||cc||cc||cc}
\toprule
\multirow{2}{*}{\textbf{Methods}} & \multicolumn{2}{c||}{\textbf{MOSI}} & \multicolumn{2}{c||}{\textbf{KS}} & \multicolumn{2}{c}{\textbf{MOSEI}} \\
 & MAE$\downarrow$ & ACC$_7$ & ACC & F1 & MAE$\downarrow$ & ACC$_7$ \\
\midrule
\midrule
\textbf{CLCR (Ours)} & \textbf{0.678} & \textbf{49.56} & \textbf{75.41} & \textbf{75.32} & \textbf{0.511} & \textbf{56.32} \\
\midrule
\midrule
\multicolumn{7}{c}{\textit{\textbf{Importance of Modality}}} \\
w/o Language   & 0.970 & 36.2 & -- & --   & 0.803 & 42.96 \\
w/o Acoustic   & 0.705 & 47.6 & 52.36 & 52.67 & 0.538 & 54.36 \\
w/o Visual     & 0.710 & 47.1 & 48.67 & 48.89 & 0.543 & 53.86 \\
only Language  & 0.727 & 46.3 & -- & --   & 0.560 & 53.06 \\
only Acoustic  & 1.020 & 33.5 & 48.67 & 48.89 & 0.853 & 40.26 \\
only Visual    & 0.950 & 35.0 & 52.36 & 52.67 & 0.783 & 41.76 \\
\midrule
\midrule
\multicolumn{7}{c}{\textit{\textbf{Importance of Critical Components}}} \\
w/o IntraCED          & 0.703 & 47.4 & 73.0 & 72.8 & 0.536 & 54.16 \\
w/o InterCAD          & 0.699 & 47.8 & 73.4 & 73.2 & 0.532 & 54.56 \\
w/o Hierarchy& 0.720 & 46.3 & 71.9 & 71.6 & 0.553 & 53.06 \\
\midrule
\midrule
\multicolumn{7}{c}{\textit{\textbf{Importance of Cross-Level Alignment}}} \\
Full Mix            & 0.743 & 45.6 & 70.3 & 70.5 & 0.585 & 52.21 \\
Partial Mix         & 0.722 & 46.4 & 71.5 & 71.6 & 0.569 & 53.76 \\
\midrule
\midrule
\multicolumn{7}{c}{\textit{\textbf{Importance of Regularization}}} \\
w/o $\mathcal{L}_{\mathrm{Inter}}$ & 0.695 & 48.1 & 74.3 & 74.0 & 0.528 & 54.86 \\
w/o $\mathcal{L}_{\mathrm{Intra}}$ & 0.690 & 48.4 & 74.6 & 74.3 & 0.523 & 55.16 \\
w/o both $\ddagger$                & 0.725 & 46.0 & 71.2 & 70.8 & 0.558 & 52.76 \\
\bottomrule
\end{tabular}
\label{tab:ablation}
\end{table}

\subsection{Ablation Studies}
\label{sec:ablation}

We perform ablations on MOSI, KS, and MOSEI to study modality configurations, critical components, cross-level alignment strategies, and regularization terms (Table~\ref{tab:ablation}).

\textbf{Importance of Modality.} To study how CLCR exploits different modalities across heterogeneous benchmarks, we vary the modality configurations. On MOSI and MOSEI, removing language causes the largest degradation, and the language-only setting performs close to full CLCR, while acoustic-only and visual-only are clearly weaker. This indicates that language is the primary sentiment carrier on these datasets, whereas audio and visual streams mainly provide complementary cues. On KS, the visual stream clearly outperforms the acoustic stream, and full CLCR consistently exceeds any single-modality or modality-dropped variant. These results suggest that CLCR adapts to the dominant modality of each benchmark and degrades gracefully when modalities are missing.

\textbf{Importance of Critical Components.} To assess the role of the semantic hierarchy and the representation modules, we ablate IntraCED, InterCAD, and the hierarchy itself. Removing either IntraCED or InterCAD leads to consistent drops on all three benchmarks, with removing IntraCED usually slightly worse, indicating that level-wise shared/private separation and controlled cross-modal exchange are crucial for reducing semantic asynchrony. Variants without the hierarchy obtain the lowest scores among these component settings, indicating that the hierarchy, IntraCED, and InterCAD are complementary and that using all of them yields more informative and stable representations than any partial design.

\textbf{Importance of Cross-Level Alignment.} To analyze the effect of cross-level semantic alignment, we conduct a perturbation study that changes only the alignment strategy while keeping all other modules fixed. We compare CLCR with two variants: in Full Mix, shallow, mid, and deep features are fully shuffled within each modality so that no level matches its true depth; in Partial Mix, exactly one level per modality remains correctly aligned and the other two are swapped. As shown in Table~\ref{tab:ablation}, Full Mix consistently performs worst, Partial Mix attains intermediate results, and CLCR achieves the lowest MAE and highest accuracy. Performance improves as cross-level alignment becomes more coherent, indicating that partial alignment is insufficient and that consistent alignment across all semantic levels is needed to fully exploit multimodal complementarity and mitigate cross-level semantic asynchrony.

\textbf{Importance of Regularization.} We also examine the regularization terms to assess the role of structural objectives. Removing either the inter-level or intra-level regularization leads to moderate but consistent degradation on benchmarks, while still keeping an advantage over variants that discard structural modules. This suggests that both regularizers help suppress redundancy and leakage and improve subspace purity. The variant without both regularizers (task loss only) gives the worst MAE and ACC$_7$ among all CLCR variants. The gap to full CLCR indicates that the semantic hierarchy, its components, and their regularization jointly support robustness and cross-dataset generalization.

\begin{figure}[ht]
\centering
\includegraphics[width=1\linewidth]{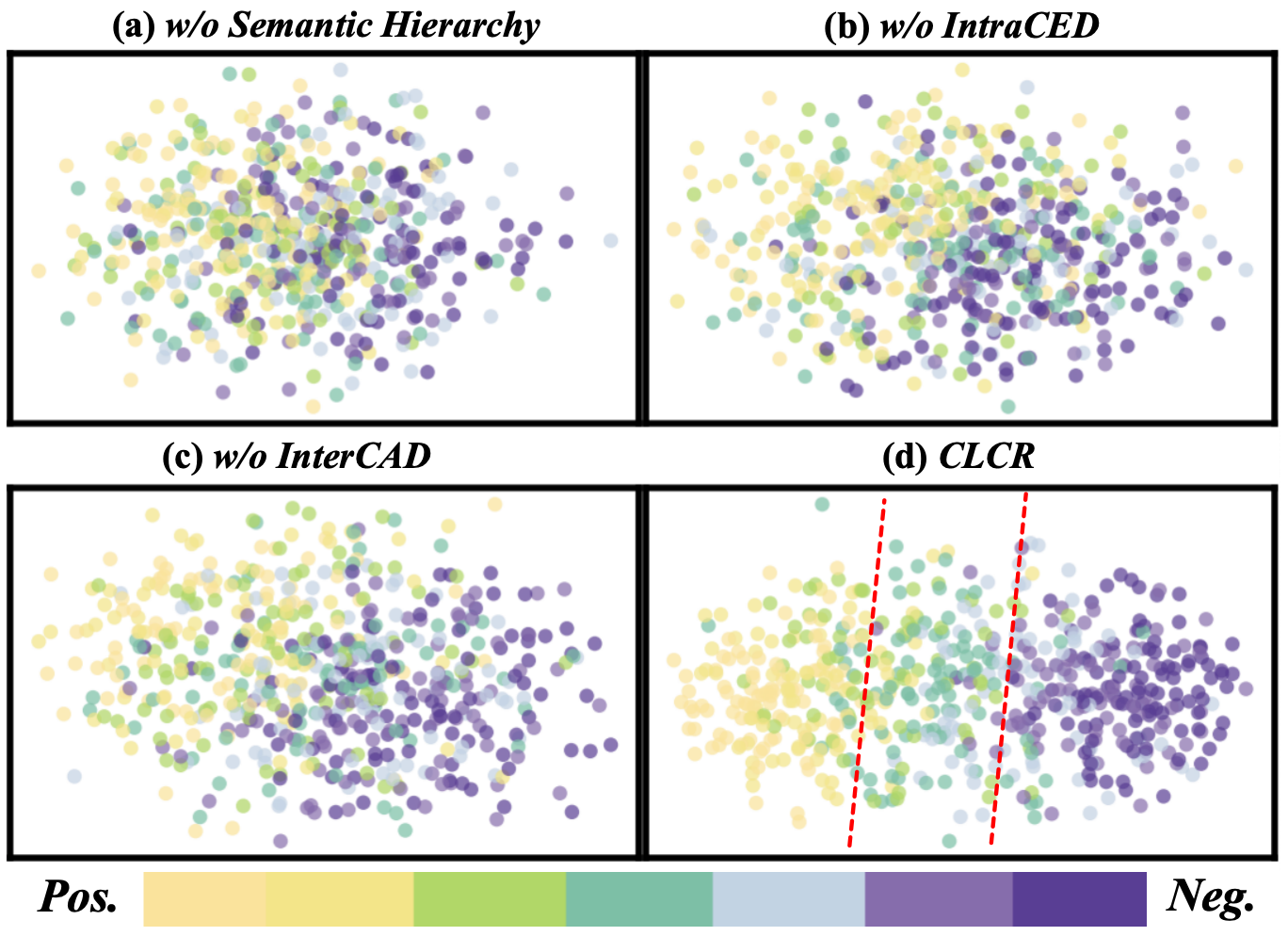}
\caption{Qualitative analysis: t-SNE of MOSI representations. CLCR yields more compact, better separated clusters than ablated variants, indicating improved semantic alignment.}
\label{fig:t-sne}
\end{figure}

\subsection{Further Analysis}

\textbf{Qualitative Analysis.} To assess the effect of the semantic hierarchy and its modules beyond scalar metrics, we visualize t-SNE~\cite{t-sne} embeddings of MOSI test features for four variants: w/o Semantic Hierarchy, w/o IntraCED, w/o InterCAD, and full CLCR (Fig.~\ref{fig:t-sne}). Without the semantic hierarchy, modality features are fused without level-wise partition or alignment and the embeddings are highly entangled, with opposite sentiments and intensity levels overlapping and no clear ordering. Without IntraCED, the distribution becomes slightly more organized but sentiment levels still mix heavily, indicating that unconstrained cross-modal exchange does not adequately reduce semantic asynchrony. Removing InterCAD yields embeddings with a clearer progression from negative to positive sentiment but with a less regular trajectory than full CLCR, reflecting the lack of explicit cross-level coordination. With full CLCR, features follow a near monotonic progression from negative to positive sentiment with limited overlap between neighbouring levels. These are consistent with the scores in Table~\ref{tab:ablation}, indicating that the semantic hierarchy together with IntraCED and InterCAD produces a more coherent and separable multimodal representation space than the ablated variants.

\begin{figure}[ht]
\centering
\includegraphics[width=0.99\linewidth]{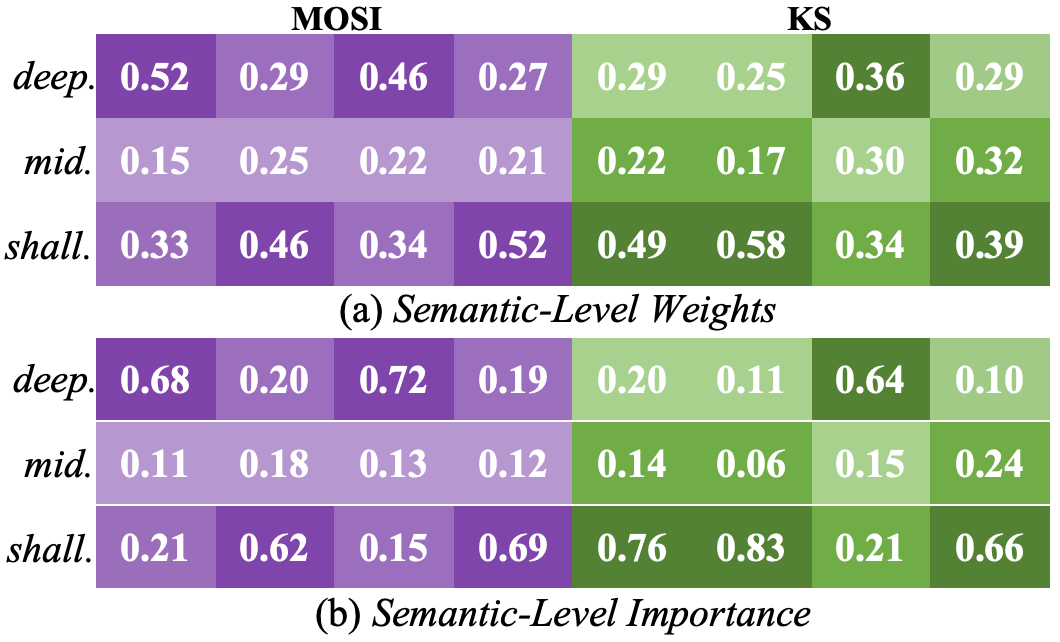}
\caption{Visualization of semantic-level importance and learned weights on MOSI and KS. The shallow, mid, and deep levels are used in a complementary rather than redundant way, supporting the necessity of the semantic hierarchy.}
\label{fig:weight+importance}
\end{figure}

\begin{figure}[ht]
\centering
\includegraphics[width=1.0\linewidth]{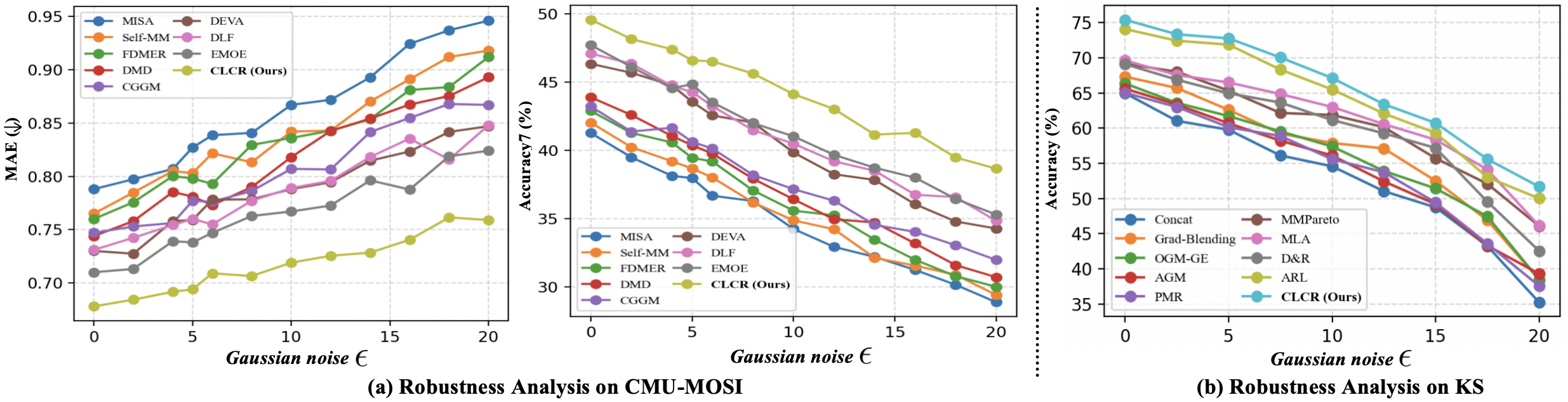}
\caption{Robustness analysis under Gaussian noise. performance curves over more than 10 random runs with different noisy data on MOSI (a) and KS (b), $\downarrow$ indicates that lower values are better.}
\label{fig:robust}
\end{figure}

\textbf{Visualization of Semantic-Level Weights and Importance.} We visualize level-wise contributions and cross-level weights on MOSI and KS to examine how CLCR utilizes its three-level semantic hierarchy, as shown in Fig.~\ref{fig:weight+importance}. For representative test samples, we estimate the contribution of each level by in turn masking shallow, mid, or deep features and observing the resulting change in prediction. We also inspect the normalized level weights produced by the cross-level aggregation module in InterCAD. On MOSI, utterances dominated by lexical content receive higher weights on deep-level features, whereas samples where intonation or facial micro-expressions are more informative rely more on the shallow level, with the mid level contributing relatively consistently across most samples. On KS, which depends more on shallow motion and acoustic events, the shallow level is strongly emphasized, while events that require contextual reasoning or object interactions increase the weights assigned to the mid and deep levels. These qualitative patterns are consistent with the results in Table~\ref{tab:ablation}: removing the semantic hierarchy or disabling cross-level aggregation leads to noticeable performance drops, while the full CLCR with all three levels and the learned level selector achieves the best performance. This supports that the three semantic levels provide complementary information and that cross-level weighting in InterCAD adapts to task semantics.

\textbf{Robustness Analysis under Gaussian Noise.} We evaluate the robustness of CLCR to noisy multimodal evidence on MOSI and KS by injecting independent zero-mean Gaussian noise with varying strengths into the input features of all modalities. As shown in Fig.~\ref{fig:robust}, the performance of all methods degrades as the noise level increases, with the drop becoming substantial at higher noise levels. Early fusion and heuristic balancing baselines exhibit the steepest degradation, whereas CLCR consistently achieves the best performance across all noise settings and shows a clearly smaller relative drop on both datasets. These results indicate that CLCR not only improves accuracy on clean inputs but also induces decision boundaries that remain stable under strong perturbations, largely due to the three-level design of IntraCED and InterCAD, which confines cross-modal exchange to shared subspaces and limits noise propagation across modalities and semantic depths.

\begin{figure}[ht]
\centering
\includegraphics[width=1.0\linewidth]{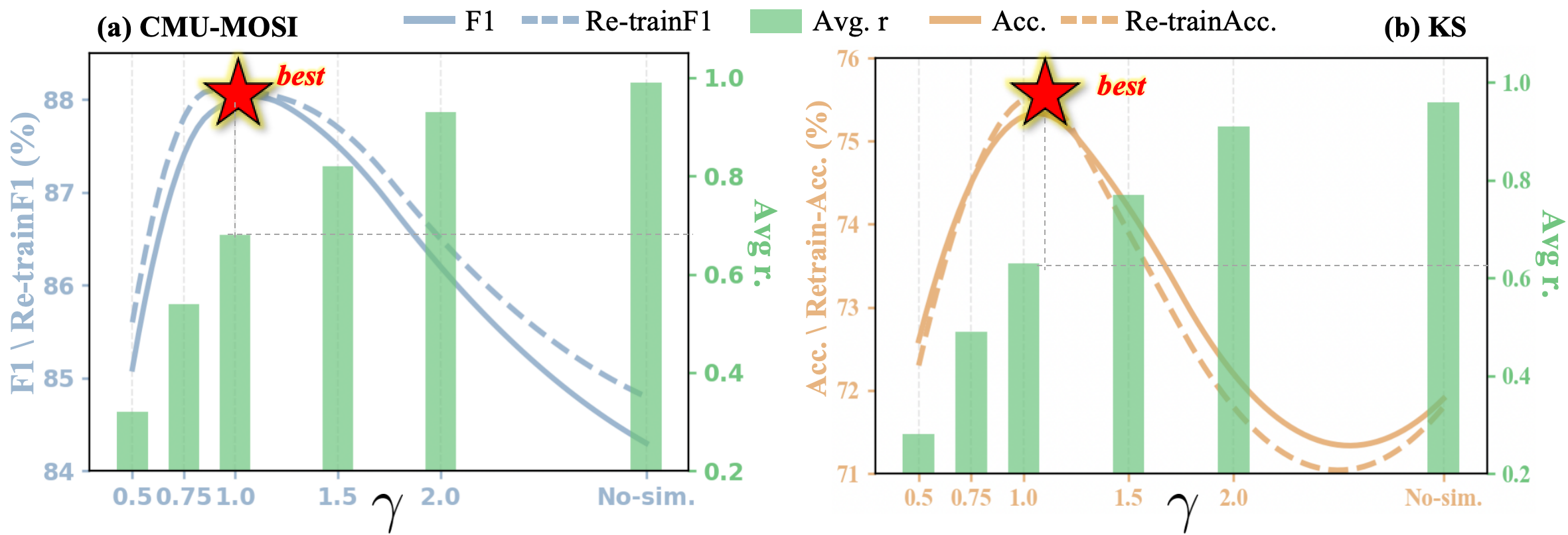}
\caption{Sparsity–performance trade-off of IntraCED token budgets. The best appears near $\gamma\!\approx\!1$ ($r\!\approx\!0.68$), indicating that a moderate sparsity level yields optimal cross-modal exchange.}
\label{fig:budget}
\end{figure}


\textbf{Analysis of IntraCED Token Budgets.} We study how token-level budgets in IntraCED control sparsity and influence model behavior via a controlled sparsity experiment. For each modality and level, we define a participation ratio $r$ as the fraction of tokens selected for cross-modal exchange, and average it over modalities and levels to obtain a single sparsity indicator. Starting from a trained CLCR model, we rescale all token budgets by a factor $\gamma$, $B_{\ell}^{(\gamma)} = \gamma B_{\ell}$, and evaluate two settings on MOSI and KS. In the first setting, we alter only $B_{\ell}$ while keeping all other parameters fixed to probe the local sensitivity of performance to sparsity around the learned budgets. In the second setting, we retrain CLCR from scratch with $B_{\ell}$ frozen at $B_{\ell}^{(\gamma)}$ under the same architecture and training protocol, testing whether alternative sparsity levels can become optimal after full adaptation. As shown in Fig.~\ref{fig:budget}, when $\gamma$ deviates from $1.0$, the participation ratio $r$ drifts towards either very sparse or nearly dense cross-modal exchange, and performance degrades in both settings and on both benchmarks. The best results consistently occur near $\gamma = 1.0$. A dense variant without simplex projection, where the participation ratio $r$ is about $0.99$ and almost all tokens participate, performs worst. These observations indicate that the token budgets in IntraCED play a crucial role in regulating how many tokens exchange information across modalities and in keeping CLCR in an empirically favorable intermediate sparsity regime.

We also assess CLCR’s robustness via sensitivity analysis of key hyperparameters (see Appendix).

%% file: sec/4_conclusion.tex
\section{Conclusion}
In this study, we address the challenge of cross-level semantic asynchrony in multimodal learning, where signals from different modalities and semantic levels are misaligned during fusion. We propose Cross-Level Co-Representation (CLCR), a novel framework that organizes each modality into a three-level semantic hierarchy. CLCR enables budgeted shared–private exchange within levels and adaptive aggregation across levels. Our experiments on audio-visual classification and multimodal sentiment analysis benchmarks demonstrate that CLCR consistently outperforms strong fusion and disentanglement baselines. This performance is further supported by ablation studies and detailed analyses of semantic-level weights and token budgets.

%% file: sec/X_suppl.tex
\clearpage
\setcounter{page}{1}
\maketitlesupplementary

\appendix

\section{Related Work}

\subsection{Multimodal Fusion and Disentangled Representations}

Early multimodal fusion approaches construct a single joint space and perform fusion in that space without distinguishing semantic levels. Tensor Fusion Network (TFN)~\cite{TFN} and Low-rank Multimodal Fusion (LMF)~\cite{LMF} model cross-modal interactions by tensor factorization and low-rank parameterization. MMBT~\cite{MMBT}, GMU-based fusion~\cite{GMU}, and language-guided architectures for multimodal transformers further exploit attention and gating to combine visual, acoustic, and textual streams. These models improve fusion quality but still project all modalities into one latent space and treat all tokens in that space as equally available for interaction. Transformer-based fusion models strengthen temporal cross-modal attention. MulT~\cite{MuLT} attends across unaligned language, vision, and audio sequences, and variants such as BPMulT~\cite{BPMulT} extend this idea to movie understanding and in-the-wild emotion recognition. They enable fine-grained token interactions but usually operate with a single interaction layer or a shallow hierarchy, so early, intermediate, and deep features are mixed without explicit control of cross-level semantics. A second line of work introduces shared and modality-specific representations, mainly for multimodal sentiment and emotion analysis. MISA~\cite{misa}, Self-MM~\cite{self-mm}, and FDMER~\cite{FDMER} separate modality-invariant and modality-specific components. DMD~\cite{dmd}, Confede~\cite{confede}, FDRL~\cite{FDRL}, and related methods refine disentanglement through contrastive objectives or decoupled distillation, while EMOE~\cite{EMOE}, DLF~\cite{DLF}, and DEVA~\cite{DEVA} design expert-style or bias-aware modules to better exploit unimodal cues. These models usually instantiate shared and private subspaces at one semantic depth and do not explicitly coordinate shared and private structure across multiple levels.

CLCR connects these directions by combining hierarchical representations with level-aware disentanglement. It organizes each modality into a three-level semantic hierarchy, factorizes each level into shared and private subspaces, and restricts cross-modal attention to a budgeted shared subspace. Inter-level aggregation is then guided by learned anchors and regularizers that align levels while suppressing redundancy, which directly targets the cross-level semantic asynchrony that is not modeled in the above fusion and disentanglement frameworks.

\begin{figure}[t]
\centering
\includegraphics[width=\linewidth]{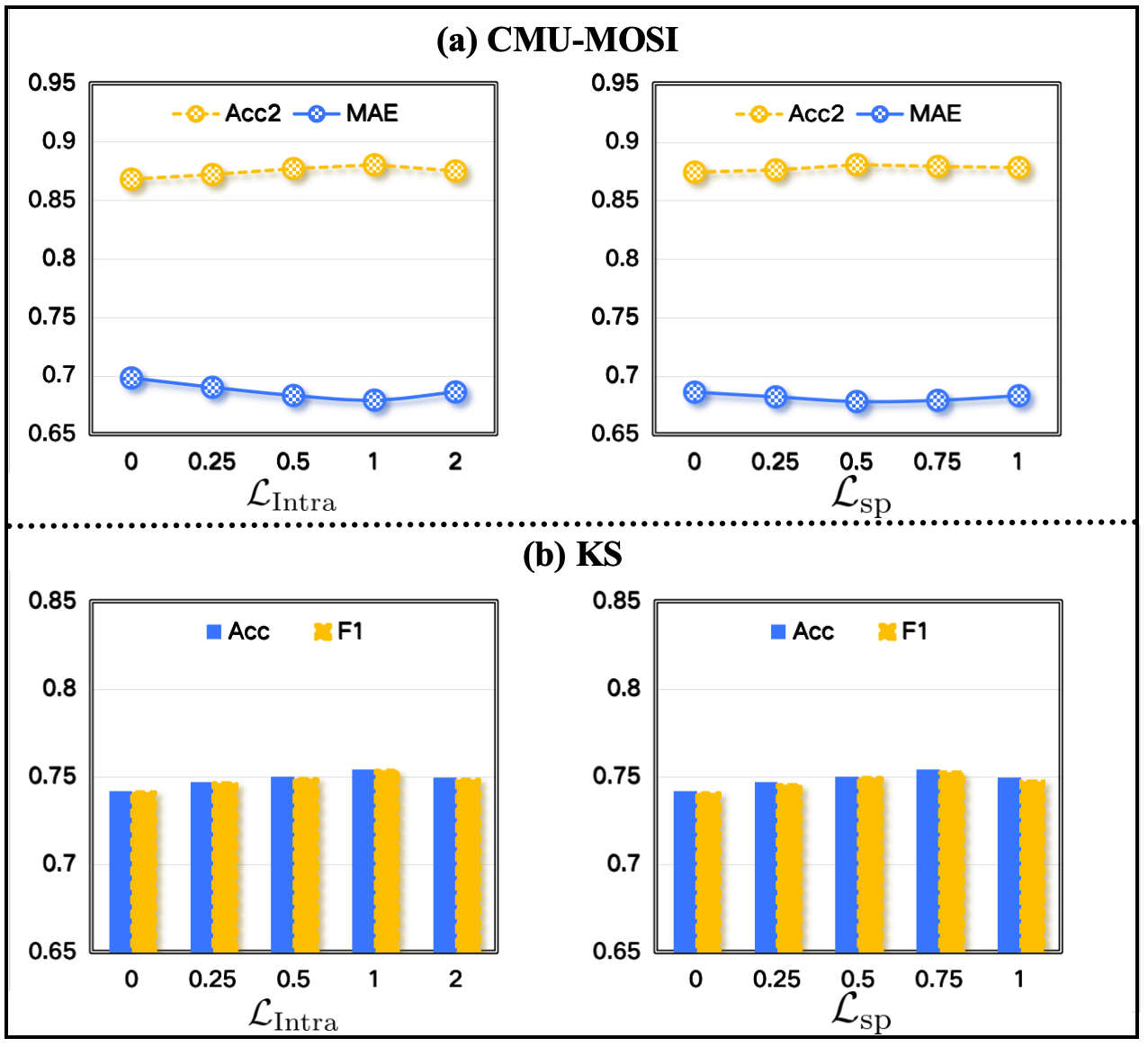}
\caption{Sensitivity of CLCR to intra-level regularization on MOSI and KS. Each plot varies the global intra-level weight $\lambda_{\mathrm{intra}}$ or the separation factor $\lambda_{\mathrm{sp}}$ in $\mathcal{L}_{\mathrm{Intra}}$, performance improves once the constraint is activated and remains stable for moderate values.}
\label{fig:sens-intra}
\end{figure}

\begin{figure*}[t]
\centering
\includegraphics[width=\linewidth]{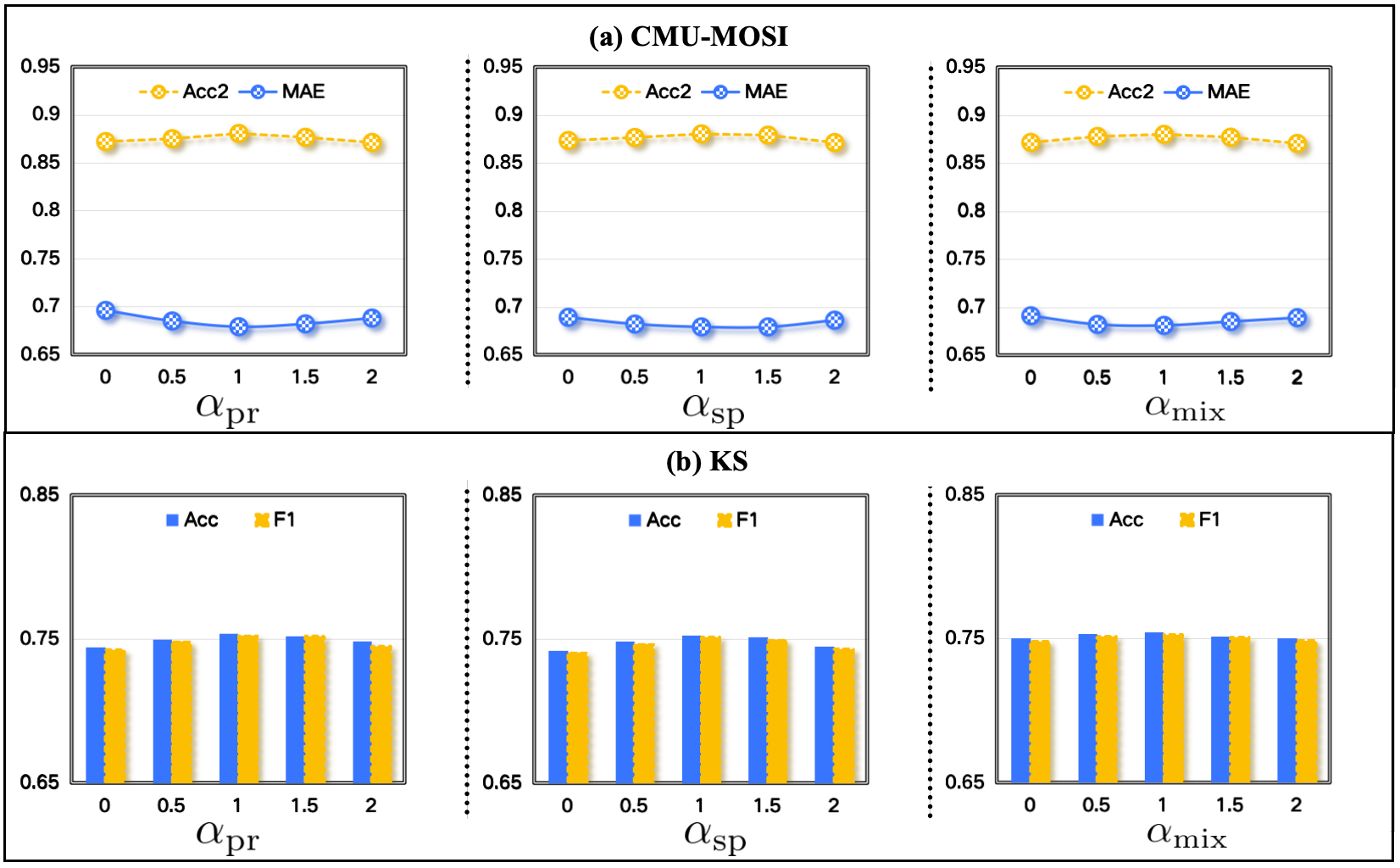}
\caption{Sensitivity of CLCR to the internal weights of the inter-level regularizer on CMU-MOSI and KS. Each plot varies one of $\alpha_{\mathrm{pr}}$, $\alpha_{\mathrm{sp}}$, or $\alpha_{\mathrm{mix}}$ while fixing the others, removing any component degrades performance, whereas weights in $[0.5,1.5]$ give similar results.}
\label{fig:sens-inter}
\end{figure*}

\subsection{Robustness Multimodal Learning}

Robust multimodal learning methods study how to keep performance stable under modality imbalance, noisy inputs, or missing data. PMR~\cite{PMR} rebalances prototypical representations to reduce the dominance of strong modalities. OGM-GE~\cite{OGM}, AGM~\cite{AGM}, CGGM~\cite{CGGM}, ARL~\cite{ARL}, DGL~\cite{DGL}, MLA~\cite{MLA}, D\&R~\cite{d2r}, and BML~\cite{BML} modulate gradients or losses using classifier feedback or multi-objective criteria so that different modalities and classes receive more balanced updates. These methods mainly act at the optimization level and do not change the internal geometry of the fused representation. Another group of works introduces robustness to low-quality or missing modalities. QMF~\cite{QMF} proposes dynamic fusion strategies for low-quality multimodal data, EAU~\cite{EAU} uses unimodal aleatoric uncertainty to guide fusion under noise. Multimodal prompting and completion approaches~\cite{PEFT} learn to handle missing inputs by predicting or compensating absent streams during inference. These methods improve robustness to imperfect observations but usually assume a single interaction space and do not explicitly control how information flows across semantic depths.

CLCR pursues robustness from a structural perspective. The semantic hierarchy, the level-wise shared and private decomposition in IntraCED, and the anchor-guided aggregation in InterCAD jointly specify where interaction is allowed, how much information is exchanged through token budgets, and how cross-level evidence is combined through correlation-based regularizers.

\section{Sensitivity Analysis}
\label{sec:sensitivity}

To understand how the regularization terms influence CLCR, we conduct a controlled sensitivity study on CMU-MOSI and KS. CMU-MOSI is a regression-style multimodal sentiment benchmark with three modalities, and KS is an audio-visual action recognition dataset. All architectural and training settings follow Section 3, and each configuration is trained with five random seeds.

\textbf{Intra-level regularization.}
We first vary the intra-level loss $\mathcal{L}_{\mathrm{Intra}}$ in Eq.7. Fig.~\ref{fig:sens-intra} reports the effect of the global weight $\lambda_{\mathrm{intra}}\in\{0,0.25,0.5,1,2\}$ and the separation factor $\lambda_{\mathrm{sp}}\in\{0.1,0.25,0.5,0.75,1\}$. When $\lambda_{\mathrm{intra}}=0$, the intra-level constraint is removed and both datasets obtain the lowest performance. Increasing $\lambda_{\mathrm{intra}}$ to around $0.5$ or $1$ reduces MAE and improves Acc$_2$ on MOSI, and raises Acc and F1 on KS. Increasing $\lambda_{\mathrm{intra}}$ to $2$ produces only minor changes, and the curves are nearly flat in the interval $[0.5,2]$. $\lambda_{\mathrm{sp}}$ shows a similar pattern: moderate values around $0.5$ achieve the best or near-best results, while smaller or larger values slightly degrade the metrics. These results indicate that CLCR benefits from intra-level regularization and remains stable once the weights fall into a moderate range.

\textbf{Inter-level weighting.}
We next analyze the internal coefficients of the inter-level loss $\mathcal{L}_{\mathrm{Inter}}$ in Eq.12. Starting from the default setting $(\alpha_{\mathrm{pr}},\alpha_{\mathrm{sp}},\alpha_{\mathrm{mix}})=(1,1,1)$ with $(\lambda_{\mathrm{inter}},\lambda_{\mathrm{intra}})=(0.1,0.1)$, we vary one coefficient at a time and keep the other two fixed. The values are $\alpha_{\mathrm{pr}}\in\{0,0.5,1,1.5,2\}$, $\alpha_{\mathrm{sp}}\in\{0.01,0.5,1,1.5,2\}$, and $\alpha_{\mathrm{mix}}\in\{0,0.5,1,1.5,2\}$. Fig.~\ref{fig:sens-inter} shows that setting any coefficient close to zero yields the highest MAE and the lowest Acc$_2$ on MOSI and the lowest Acc and F1 on KS, which means that removing the corresponding term harms performance. When the weight moves from zero into the interval $[0.5,1.5]$, all metrics improve and then vary only slightly within this interval. The balanced choice $(1,1,1)$ lies near the center of this plateau on both datasets and therefore serves as a convenient default in all experiments.

Taken together, these sensitivity curves show that the regularizers play a useful role without introducing fragile hyperparameter dependencies. The main gains of CLCR come from its cross-level architecture, while the regularization terms act as structural guidance whose weights can be chosen from a broad and stable range.

%% file: main.bib
@String(CVPR= {IEEE Conf. Comput. Vis. Pattern Recog.})

@String(ICCV= {Int. Conf. Comput. Vis.})

@String(ICASSP=	{ICASSP})

@String(ICLR = {Int. Conf. Learn. Represent.})

@String(AAAI = {AAAI})

@String(CVPR  = {CVPR})

@String(ICCV  = {ICCV})

@String(ICLR  = {ICLR})

@article{BPMulT,
title={Automatic movie genre classification \& emotion recognition via a BiProjection Multimodal Transformer},
author={Moreno-Galván, Diego Aarón and others},
journal={Information Fusion},
volume={102},
pages={102641},
year={2024},
publisher={Elsevier}
}

@inproceedings{EAU,
title={Embracing Unimodal Aleatoric Uncertainty for Robust Multimodal Fusion},
author={Gao, Xun and Jiang, Xun and others},
booktitle={Proceedings of the IEEE/CVF Conference on Computer Vision and Pattern Recognition (CVPR)},
year={2024}
}

@inproceedings{MLA,
title={Multimodal Representation Learning by Alternating Unimodal Adaptation},
author={Zhang, Xiaohui and Yoon, Jaehong and Bansal, Mohit and Yao, Huaxiu},
booktitle={Proceedings of the IEEE/CVF Conference on Computer Vision and Pattern Recognition (CVPR)},
year={2024}
}

@inproceedings{BML,
title={Balancing Multimodal Learning with Classifier-guided Gradient Modulation},
author={Zhou, Zhao and others},
booktitle={Advances in Neural Information Processing Systems (NeurIPS)},
year={2024}
}

@inproceedings{QMF,
title={Provable Dynamic Fusion for Low-Quality Multimodal Data},
author={Zhang, Qingyang and Wu, Haitao and Zhang, Changqing and Hu, Qinghua and Fu, Huazhu and Zhou, Joey Tianyi and Peng, Xi},
booktitle={Proceedings of the 40th International Conference on Machine Learning (ICML)},
year={2023}
}

@inproceedings{Han2021,
title={Improving Multimodal Fusion with Hierarchical Mutual Information Maximization for Multimodal Sentiment Analysis},
author={Han, Wei and Chen, Hui and Siong, Siau-Cheng},
booktitle={Proceedings of the 2021 Conference on Empirical Methods in Natural Language Processing (EMNLP)},
pages={9191--9201},
year={2021}
}

@inproceedings{Zhang2021,
title={Hierarchical Cross-Modality Semantic Correlation Learning Model for Multimodal Summarization},
author={Zhang, Litian and Zhang, Xiaoming and Pan, Junshu and Huang, Feiran},
booktitle={Proceedings of the 35th AAAI Conference on Artificial Intelligence},
year={2021}
}

@inproceedings{tishby2000information,
  title={The information bottleneck method},
  author={Tishby, Naftali and Pereira, Fernando C and Bialek, William},
  booktitle={Proceedings of the 37th annual Allerton conference on communication, control and computing},
  volume={39},
  pages={368--377},
  year={2000}
}

@article{MENG-MGJR,
title = {Multi-grained teacher–student joint representation learning for surface defect classification},
journal = {Journal of Industrial Information Integration},
volume = {48},
pages = {100958},
year = {2025},
author = {Chunlei Meng and Jiacheng Yang and Wei Lin and Linqiang Hu and Bowen Liu and Zhuo Zou and LiDa Xu and Zhongxue Gan and Chun Ouyang}
}

@inproceedings{GMU,
  title     = {Gated Multimodal Units for Information Fusion},
  author    = {Arevalo, Juli{\'a}n and Solorio, Thamar and Montes, Manuel and Gonzalez, Fabio A.},
  booktitle = {International Conference on Learning Representations (ICLR)},
  year      = {2017}
}

@inproceedings{MMBT,
  title     = {Supervised Multimodal Bitransformers for Classifying Images and Text},
  author    = {Kiela, Douwe and Bulian, Jannis and Clark, Christopher and Mohan, Armand and Wang, Danqing and Scialom, Thomas and Williams, Adina and Weston, Jason},
  booktitle = {Proceedings of the Conference on Empirical Methods in Natural Language Processing (EMNLP)},
  pages     = {9268--9280},
  year      = {2020}
}

@inproceedings{PEFT,
  title     = {Missing Modality Prediction for Unpaired Multimodal Learning via Joint Embedding of Unimodal Models},
  author    = {Yang, Yingjie and Zhang, Hexin and Han, Kai and Wu, Tong and Song, Jizhe and Liu, Jing and Liu, Ziwei},
  booktitle = {Proceedings of the IEEE/CVF International Conference on Computer Vision (ICCV)},
  pages     = {1275--1285},
  year      = {2023}
}

@inproceedings{IB,
  title={Information bottleneck for Gaussian variables},
  author={Chechik, Gal and Globerson, Amir and Tishby, Naftali and Weiss, Yair},
  booktitle = {Proceedings of the Advances in Neural Information Processing Systems (NeurIPS)},
  volume={16},
  year={2003}
}

@article{t-sne,
  title={Visualizing data using t-SNE.},
  author={Van der Maaten, Laurens and Hinton, Geoffrey},
  journal={Journal of machine learning research},
  volume={9},
  number={11},
  year={2008}
}

@inproceedings{misa,
  title={Misa: Modality-invariant and-specific representations for multimodal sentiment analysis},
  author={Hazarika, Devamanyu and Zimmermann, Roger and Poria, Soujanya},
  booktitle={Proceedings of the 28th ACM international conference on multimedia},
  pages={1122--1131},
  year={2020}
}

@inproceedings{FDRL,
  title={Fine-grained disentangled representation learning for multimodal emotion recognition},
  author={Sun, Haoqin and Zhao, Shiwan and Wang, Xuechen and Zeng, Wenjia and Chen, Yong and Qin, Yong},
  booktitle={ICASSP 2024-2024 IEEE International Conference on Acoustics, Speech and Signal Processing (ICASSP)},
  pages={11051--11055},
  year={2024},
  organization={IEEE}
}

@inproceedings{FDMER,
  title={Disentangled representation learning for multimodal emotion recognition},
  author={Yang, Dingkang and Huang, Shuai and Kuang, Haopeng and Du, Yangtao and Zhang, Lihua},
  booktitle={Proceedings of the 30th ACM international conference on multimedia},
  pages={1642--1651},
  year={2022}
}

@inproceedings{dmd,
  title={Decoupled multimodal distilling for emotion recognition},
  author={Li, Yong and Wang, Yuanzhi and Cui, Zhen},
  booktitle={Proceedings of the IEEE/CVF conference on computer vision and pattern recognition},
  pages={6631--6640},
  year={2023}
}

@inproceedings{confede,
  title={Confede: Contrastive feature decomposition for multimodal sentiment analysis},
  author={Yang, Jiuding and Yu, Yakun and Niu, Di and Guo, Weidong and Xu, Yu},
  booktitle={Proceedings of the 61st Annual Meeting of the Association for Computational Linguistics},
  pages={7617--7630},
  year={2023}
}

@article{CGGM,
  title={Classifier-guided gradient modulation for enhanced multimodal learning},
  author={Guo, Zirun and Jin, Tao and Chen, Jingyuan and Zhao, Zhou},
  journal={Advances in Neural Information Processing Systems},
  volume={37},
  pages={133328--133344},
  year={2024}
}

@inproceedings{d2r,
  title={D2r: Dual-branch dynamic routing network for multimodal sentiment detection},
  author={Chen, Yifan and Li, Kuntao and Mai, Weixing and Wu, Qiaofeng and Xue, Yun and Li, Fenghuan},
  booktitle={Proceedings of the 2024 Conference on Empirical Methods in Natural Language Processing},
  pages={3536--3547},
  year={2024}
}

@article{TCN,
  title={An empirical evaluation of generic convolutional and recurrent networks for sequence modeling},
  author={Bai, Shaojie and Kolter, J Zico and Koltun, Vladlen},
  journal={arXiv preprint arXiv:1803.01271},
  year={2018}
}

@article{Cmu-mosi,
  title={Multimodal sentiment intensity analysis in videos: Facial gestures and verbal messages},
  author={Zadeh, Amir and Zellers, Rowan and Pincus, Eli and Morency, Louis-Philippe},
  journal={IEEE Intelligent Systems},
  pages={82--88},
  year={2016}
}

@inproceedings{Cmu-mosei,
  title={Multimodal language analysis in the wild: Cmu-mosei dataset and interpretable dynamic fusion graph},
  author={Zadeh, AmirAli Bagher and Liang, Paul Pu and Poria, Soujanya and Cambria, Erik and Morency, Louis-Philippe},
  booktitle={Proceedings of the 56th Annual Meeting of the Association for Computational Linguistics},
  pages={2236--2246},
  year={2018}
}

@inproceedings{EMOE,
  title={EMOE: Modality-Specific Enhanced Dynamic Emotion Experts},
  author={Fang, Yiyang and Huang, Wenke and Wan, Guancheng and Su, Kehua and Ye, Mang},
  booktitle={Proceedings of the Computer Vision and Pattern Recognition Conference},
  pages={14314--14324},
  year={2025}
}

@inproceedings{DLF,
  title={DLF: Disentangled-language-focused multimodal sentiment analysis},
  author={Wang, Pan and Zhou, Qiang and Wu, Yawen and Chen, Tianlong and Hu, Jingtong},
  booktitle={Proceedings of the AAAI Conference on Artificial Intelligence},
  volume={39},
  pages={21180--21188},
  year={2025}
}

@article{TSDA,
      title={Temporal-Spatial Decouple before Act: Disentangled Representation Learning for Multimodal Sentiment Analysis}, 
      author={Chunlei Meng and Ziyang Zhou and Lucas He and Xiaojing Du and Chun Ouyang and Zhongxue Gan},
      journal={arXiv preprint arXiv:2601.13659},
      year={2026}
}

@inproceedings{CF-ViT,
  author={Meng, Chunlei and Lin, Wei and Yang, Jiacheng and Liu, Yi and Zhang, Hongda and Chen, Yuning and Liu, Bowen and Zhou, Ziqin and Ouyang, Chun and Gan, Zhongxue and Wu, Dunzhao and Nie, Zhihua},
  booktitle={2025 IEEE International Conference on Systems, Man, and Cybernetics (SMC)}, 
  title={CF-ViT: Cross-Feature Vision Transformer for Improving Feature Learning on Tiny Datasets}, 
  year={2025},
  pages={6919-6926}
}

@inproceedings{DEVA,
  title={Enriching multimodal sentiment analysis through textual emotional descriptions of visual-audio content},
  author={Wu, Sheng and He, Dongxiao and Wang, Xiaobao and Wang, Longbiao and Dang, Jianwu},
  booktitle={Proceedings of the AAAI Conference on Artificial Intelligence},
  volume={39},
  pages={1601--1609},
  year={2025}
}

@inproceedings{Semi-IIN,
  title={Semi-IIN: Semi-supervised Intra-inter modal Interaction Learning Network for Multimodal Sentiment Analysis},
  author={Lin, Jinhao and Wang, Yifei and Xu, Yanwu and Liu, Qi},
  booktitle={Proceedings of the AAAI Conference on Artificial Intelligence},
  volume={39},
  pages={1411--1419},
  year={2025}
}

@inproceedings{MuLT,
  title={Multimodal transformer for unaligned multimodal language sequences},
  author={Tsai, Yao-Hung Hubert and Bai, Shaojie and Liang, Paul Pu and Kolter, J Zico and Morency, Louis-Philippe and Salakhutdinov, Ruslan},
  booktitle={Proceedings of the conference. Association for computational linguistics. Meeting},
  pages={6558},
  year={2019}
}

@inproceedings{TFN,
    title = {Tensor Fusion Network for Multimodal Sentiment Analysis},
    author = {Zadeh, Amir  and Chen, Minghai  and
      Poria, Soujanya  and
      Cambria, Erik  and
      Morency, Louis-Philippe},
    booktitle = {Proceedings of the 2017 Conference on Empirical Methods in Natural Language Processing},
    pages = {1103--1114}, 
  year = {2017}
}

@article{LMF,
  title={Efficient low-rank multimodal fusion with modality-specific factors},
  author={Liu, Zhun and Shen, Ying and Lakshminarasimhan, Varun Bharadhwaj and Liang, Paul Pu and Zadeh, Amir and Morency, Louis-Philippe},
  journal={arXiv preprint arXiv:1806.00064},
  year={2018}
}

@inproceedings{self-mm,
  title={Learning modality-specific representations with self-supervised multi-task learning for multimodal sentiment analysis},
  author={Yu, Wenmeng and Xu, Hua and Yuan, Ziqi and Wu, Jiele},
  booktitle={Proceedings of the AAAI conference on artificial intelligence},
  pages={10790--10797},
  year={2021}
}

@inproceedings{MAG-BERT,
  author       = {Wasifur Rahman and Md. Kamrul Hasan and Sangwu Lee and AmirAli Bagher Zadeh and Chengfeng Mao and Louis{-}Philippe Morency and Mohammed E. Hoque},
  title        = {Integrating Multimodal Information in Large Pretrained Transformers},
  booktitle    = {Proceedings of the 58th Annual Meeting of the Association for Computational Linguistics},
  pages        = {2359--2369},
  year         = {2020}
}

@article{cta-net,
  title={CTA-Net: A CNN-Transformer Aggregation Network for Improving Multi-Scale Feature Extraction},
  author={Meng, Chunlei and Yang, Jiacheng and Lin, Wei and Liu, Bowen and Zhang, Hongda and Gan, Zhongxue and others},
  journal={arXiv preprint arXiv:2410.11428},
  year={2024}
}

@ARTICLE{rts-vit,
  author={Meng, Chunlei and Lin, Wei and Liu, Bowen and Zhang, Hongda and Gan, Zhongxue and Ouyang, Chun},
  journal={IEEE Journal of Biomedical and Health Informatics}, 
  title={RTS-ViT: Real-Time Share Vision Transformer for Image Classification}, 
  year={2025},
  volume={29},
  number={5},
  pages={3576-3586},
 }

@inproceedings{ARL,
  title={Improving Multimodal Learning via Imbalanced Learning},
  author={Wei, Shicai and Luo, Chunbo and Luo, Yang},
  booktitle={Proceedings of the IEEE/CVF International Conference on Computer Vision},
  pages={2250--2259},
  year={2025}
}

@inproceedings{DGL,
  title={Boosting Multimodal Learning via Disentangled Gradient Learning},
  author={Wei, Shicai and Luo, Chunbo and Luo, Yang},
  booktitle={Proceedings of the IEEE/CVF International Conference on Computer Vision},
  year={2025}
}

@article{mmpareto,
  title={Mmpareto: Boosting multimodal learning with innocent unimodal assistance},
  author={Wei, Yake and Hu, Di},
  journal={arXiv preprint arXiv:2405.17730},
  year={2024}
}

@inproceedings{2025-low,
    title = {Low-Rank Interconnected Adaptation across Layers},
    author = {Zhong, Yibo  and
      Zhao, Jinman  and
      Zhou, Yao},
    booktitle = {Findings of the Association for Computational Linguistics: ACL 2025},
   pages = {17005--17029},
    year = {2025},
}

@inproceedings{PMR,
  title={Pmr: Prototypical modal rebalance for multimodal learning},
  author={Fan, Yunfeng and Xu, Wenchao and Wang, Haozhao and Wang, Junxiao and Guo, Song},
  booktitle={Proceedings of the IEEE/CVF Conference on Computer Vision and Pattern Recognition},
  pages={20029--20038},
  year={2023}
}

@inproceedings{AGM,
  title={Boosting multi-modal model performance with adaptive gradient modulation},
  author={Li, Hong and Li, Xingyu and Hu, Pengbo and Lei, Yinuo and Li, Chunxiao and Zhou, Yi},
  booktitle={Proceedings of the IEEE/CVF International Conference on Computer Vision},
  pages={22214--22224},
  year={2023}
}

@inproceedings{OGM,
  title={Balanced multimodal learning via on-the-fly gradient modulation},
  author={Peng, Xiaokang and Wei, Yake and Deng, Andong and Wang, Dong and Hu, Di},
  booktitle={Proceedings of the IEEE/CVF conference on computer vision and pattern recognition},
  pages={8238--8247},
  year={2022}
}

@inproceedings{KS,
  title={Look, listen and learn},
  author={Arandjelovic, Relja and Zisserman, Andrew},
  booktitle={Proceedings of the IEEE international conference on computer vision},
  pages={609--617},
  year={2017}
}

@article{crema-D,
  title={Crema-d: Crowd-sourced emotional multimodal actors dataset},
  author={Cao, Houwei and Cooper, David G and Keutmann, Michael K and Gur, Ruben C and Nenkova, Ani and Verma, Ragini},
  journal={IEEE transactions on affective computing},
  volume={5},
  number={4},
  pages={377--390},
  year={2014}
}

@article{ucf101,
  title={Ucf101: A dataset of 101 human actions classes from videos in the wild},
  author={Soomro, Khurram and Zamir, Amir Roshan and Shah, Mubarak},
  journal={arXiv preprint arXiv:1212.0402},
  year={2012}
}
